\pdfoutput=1

\documentclass[11pt]{article}

\usepackage[]{acl}

\usepackage{times}
\usepackage{latexsym}

\usepackage[T1]{fontenc}

\usepackage[utf8]{inputenc}

\usepackage{microtype}

\usepackage{inconsolata}
\usepackage{booktabs} 
\usepackage{pifont}
\newcommand{\CHECK}{\textcolor{green}{\ding{51}}} 
\newcommand{\CROSS}{\textcolor{red}{\ding{55}}} 
\usepackage{graphicx} 
\usepackage{multirow}
\usepackage{multicol}
\usepackage{arydshln}
\usepackage{color}
\usepackage{amssymb}

\title{emotion2vec: Self-Supervised Pre-Training \\ for Speech Emotion Representation}

\author{
Ziyang Ma\textsuperscript{1},
Zhisheng Zheng\textsuperscript{1},
Jiaxin Ye\textsuperscript{2},
Jinchao Li\textsuperscript{3}, \\
\textbf{
Zhifu Gao\textsuperscript{4},
Shiliang Zhang\textsuperscript{4},
Xie Chen\textsuperscript{1}\footnotemark[2]} \\
$^1$ Shanghai Jiao Tong University, $^2$ Fudan University, \\
$^3$ The Chinese University of Hong Kong, $^4$ Alibaba \\ 
}

\begin{document}
\maketitle

\renewcommand{\thefootnote}{\fnsymbol{footnote}}
\footnotetext[2]{Corresponding author}
\renewcommand{\thefootnote}{\arabic{footnote}}

\begin{abstract}
We propose emotion2vec, a universal speech emotion representation model. 
emotion2vec is pre-trained on open-source unlabeled emotion data through self-supervised online distillation, combining utterance-level loss and frame-level loss during pre-training. 
emotion2vec outperforms state-of-the-art pre-trained universal models and emotion specialist models by only training linear layers for the speech emotion recognition task on the mainstream IEMOCAP dataset. 
In addition, emotion2vec shows consistent improvements among 10 different languages of speech emotion recognition datasets. 
emotion2vec also shows excellent results on other emotion tasks, such as song emotion recognition, emotion prediction in conversation, and sentiment analysis. 
Comparison experiments, ablation experiments, and visualization comprehensively demonstrate the universal capability of the proposed emotion2vec. 
To the best of our knowledge, emotion2vec is the first universal representation model in various emotion-related tasks, filling a gap in the field.~\footnote{Code, checkpoints, and extracted features are available at \url{https://github.com/ddlBoJack/emotion2vec}} 

\end{abstract}

\section{Introduction}
Extracting emotional representation from speech is an essential step of various emotional tasks such as speech emotion recognition (SER) and sentiment analysis. 
Traditional methods employ Filter Banks (FBanks) or Mel Frequency Cepstrum Coefficients (MFCCs) as speech features. These features are not rich in semantic information, resulting in limited performance on emotional tasks. 
Popular methods utilize features extracted from speech-based self-supervised learning (SSL) pre-trained models, leading to a significant performance improvement. 

One potential challenge blocking further performance improvement is that these SSL models are not entirely suitable for emotional tasks. 
\citet{wang2021fine} explore no fine-tuning, partial fine-tuning, and entire fine-tuning with some SSL models for SER on the IEMOCAP dataset~\cite{busso2008iemocap}, and give some empirical conclusions. 
While this is an ad-hoc solution, on the one hand, fine-tuning SSL models requires a large computational cost, on the other hand, these conclusions may be data-specific or model-constrained.
Recently, \citet{chen2023vesper} proposed an SER model named Vesper, which is obtained by model distillation from WavLM-large~\cite{chen2022wavlm} with emotion data. 
Vesper is designed to perform the SER task, whose universal representation capability still needs to be demonstrated. 
Accordingly, a universal speech-based emotion representation model is urgently needed in the field. 

Here we propose emotion2vec, a universal emotion representation model that can be used to extract speech features for diverse emotion tasks. 
Self-supervised pre-training is performed on 262 hours of open-source emotion data with an online distillation paradigm to obtain emotion2vec. 
Considering that both whole-play information and local details convey emotion, we propose a pre-training strategy combining utterance-level loss and frame-level loss. 
On the mainstream IEMOCAP dataset, the downstream linear model trained with features extracted from emotion2vec outperforms all the mainstream SSL models and the latest specialist models. 
emotion2vec is tested on 13 datasets including 10 languages, and the results show that emotion2vec exhibits language generalization ability. 
Moreover, in addition to the SER task, we also experimented with emotion2vec features on song emotion recognition, emotion prediction in conversation, and sentiment analysis. 
The results indicate that emotion2vec has excellent task generalization ability. 
Extensive ablation experiments and visualization analysis demonstrate the effectiveness of our pre-training methods and the versatility of the proposed emotion2vec model. 

\section{Related Work}

\subsection{Speech-based SSL}
Self-supervised learning has achieved remarkable success in the field of representation learning, showcasing its efficacy across natural language processing~\cite{kenton2019bert, liu2019roberta, radford2019language, brown2020language}, computer vision~\cite{grill2020bootstrap, he2020momentum, bao2021beit, he2022masked}, as well as speech processing~\cite{baevski2020wav2vec, hsu2021hubert, chen2022wavlm, baevski2022data2vec}. 
For speech representation learning, all SSL models can be classified into two categories according to the self-supervised targets utilized during pre-training~\cite{ma2022mt4ssl}: \textbf{1)} Offline targets. \textbf{2)} Online targets. 
Models employing offline targets often require a well-trained teacher model before the pre-training stage, to extract self-supervised targets. 
Representative models of this type are HuBERT~\cite{hsu2021hubert}, WavLM~\cite{chen2022wavlm} using K-means targets, and PBERT~\cite{wang2022supervision}, MonoBERT\&PolyBERT~\cite{ma2023pushing} using phoneme-based targets. 
Models using online targets do not need a pre-trained teacher model in advance, while the teacher models are constantly updated during the pre-training phase, with an online distillation paradigm.
Representative models of this type are data2vec~\cite{baevski2022data2vec}, data2vec 2.0~\cite{baevski2023efficient} using frame-level mask language model (MLM) loss, and CA-DINO~\cite{han2023self} using utterance-level cross-entropy loss. 
emotion2vec is pre-trained combining both utterance-level loss and frame-level loss, leading to a superior speech emotion representation model. 

\subsection{Speech Emotion Representation}
We present the first universal speech emotion representation model, whereas most of the previous works directly employ speech pre-training models~\cite{pepino2021emotion, li2023exploration}, or fine-tune the pre-training models on their specific emotional data with specific emotional tasks (mostly SER)~\cite{morais2022speech, chen2023exploring}, to extract speech emotion representation. 
A series of works investigate the SER performance of wav2vec 2.0~\cite{wang2021fine}, HuBERT~\cite{wang2021fine}, as well as WavLM~\cite{ioannides2023towards}, either fine-tuning or not. 
A recent work~\cite{ma2023leveraging} found that data2vec features also have a good representation ability in the SER task. 
For speech emotion representation in other emotion tasks, such as multimodal emotion recognition, popular practice~\cite{li2022context} is similar to what is mentioned above. 

\section{Methods}

Here we mainly introduce the self-supervised pre-training method of the proposed emotion2vec, for which the core is to train the model with \textbf{Utterance-level Loss} and \textbf{Frame-level Loss} using \textbf{Online Distillation} paradigm. 

\begin{figure*}[htbp]
  \centering
  \includegraphics[width=1\textwidth]{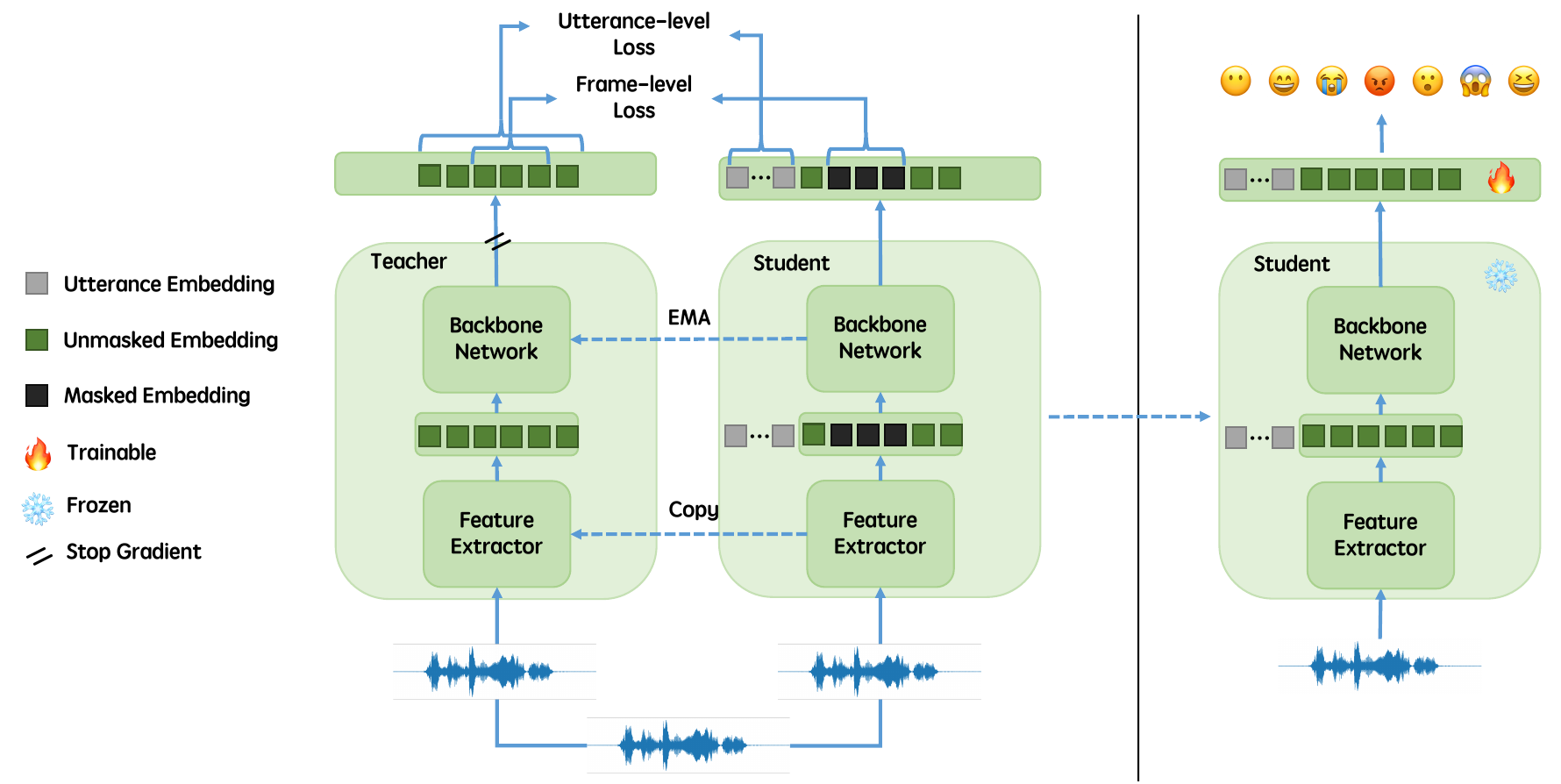}
  \caption{The overall framework of emotion2vec. During the pre-training phase, emotion2vec conducts online distillation with a teacher network and a student network. When a specific downstream task is performed, emotion2vec is frozen and a lightweight downstream model is trained.}
  \label{fig:emotion2vec}
\end{figure*}

\subsection{Model Pipeline} 
As shown in Figure~\ref{fig:emotion2vec}, emotion2vec contains two networks in the pre-training phase, which are the teacher network $\mathcal{T}$ and the student network $\mathcal{S}$. 
Both models share the same model architecture, including a feature extractor $\mathcal{F}$ composed of multi-layer convolutional neural networks and a backbone network $\mathcal{B}$ composed of multi-layer Transformers. These modules can be configured with different architectures, which will be described in Section~\ref{sec:initial_model}.
Given a raw audio utterance $X = [x_1, \cdots, x_{N_x}]$, the Teacher $\mathcal{T}$ and the Student $\mathcal{S}$ respectively utilize feature extractors $\mathcal{F}^{\mathcal{T}}$ and $\mathcal{F}^{\mathcal{S}}$ to obtain the downsampled features $Z = [z_1, \cdots, z_{N_z}]$, which can be written as: 
\begin{equation}
Z^{\mathcal{T}} = \mathcal{F}^{\mathcal{T}}(X),
\end{equation}
\begin{equation}
Z^{\mathcal{S}} = \mathcal{F}^{\mathcal{S}}(X).
\end{equation}
For the teacher network $\mathcal{T}$, the downsampled features $Z^{\mathcal{T}}$ are directly fed into the backbone network $\mathcal{B}^{\mathcal{T}}$.
For the student network $\mathcal{S}$, the downsampled features $Z^{\mathcal{S}}$ are masked $l$ consecutive frames with probability $p$ for each frame as the start.  Then learnable utterance embedding $U = [u_1, \cdots, u_{N_u}]$ is placed in the front before being fed into the backbone network $\mathcal{B}^{\mathcal{S}}$. 
The formula can be written as follows: 
\begin{equation}
Y^{\mathcal{T}} = \frac{1}{k} \sum_{i=1}^{k} \mathcal{B}_i^{\mathcal{T}}(Z^{\mathcal{T}}),
\end{equation}
\begin{equation}
U^{\mathcal{S}};Y^{\mathcal{S}} = \mathcal{B}^{\mathcal{S}}(U;Mask(Z^{\mathcal{S}})),
\end{equation}
where $Y^{\mathcal{T}}$ is the average of the output embedding of the top $k$ layer Transformer Block $\mathcal{B}_i^{\mathcal{T}}$. Utterance-level output embedding $U^{\mathcal{S}}$ and frame-level output embedding $Y^{\mathcal{S}}$ are the outputs of the student backbone network $\mathcal{B}^{\mathcal{S}}$. 
$Mask$ is the applying mask operation. 
$Y^{\mathcal{T}}$, $Y^{\mathcal{S}}$ and $U^{\mathcal{S}}$ are the same in the hidden layer dimensions, where $Y^{\mathcal{T}}$ and $Y^{\mathcal{S}}$ have the same $N_z$ temporal dimensions, while $U^{\mathcal{S}}$ has $N_u$ temporal dimensions, respectively.

\subsection{Utterance-level Loss} 
\label{sec:utterance-level loss}
\begin{figure}[htbp]
  \centering
  \includegraphics[width=1\linewidth]{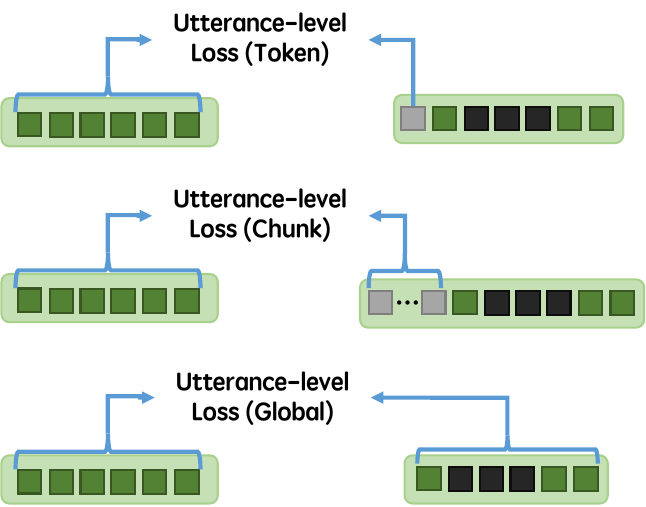}
  \caption{Different ways to compute utterance-level loss in emotion2vec pre-training. }
  \label{fig:utterance-level-loss}
\end{figure}

Utterance-level loss constructs an utterance-level pretext task to learn the global emotion. 
We use mean squared error (MSE) to calculate the loss, which can be written as:
\begin{equation}
L_{Utt} = (\bar{Y}^{\mathcal{T}} - \bar{U}^{\mathcal{S}})^2,
\end{equation}
where
\begin{equation}
\bar{Y}^{\mathcal{T}} = \frac{1}{N_z} \sum_{i=1}^{N_z} Y_i^{\mathcal{T}},
\end{equation}
\begin{equation}
\bar{U}^{\mathcal{S}} = \frac{1}{N_u} \sum_{i=1}^{N_u} U_i^{\mathcal{S}},
\end{equation}
which means that utterance-level loss $L_{Utt}$ is computed by temporal pooling results of $Y^{\mathcal{T}}$ and $U^{\mathcal{S}}$. 
Here we propose three ways to compute utterance-level loss, which we call \textbf{token embedding}, \textbf{chunk embedding}, and \textbf{global embedding}, as shown in Figure~\ref{fig:utterance-level-loss}. 
\paragraph{Token Embedding} Token embedding employs a single token to represent global emotion information encoded by the student network $\mathcal{S}$. 
More explicitly, we set $N_u$ to 1 in the learnable utterance embedding $U = [u_1, \cdots, u_{N_u}]$. 
\paragraph{Chunk Embedding} Chunk embedding employs multiple tokens to represent global emotion information. In this case, more global information can be aggregated within the chunk. 
\paragraph{Global Embedding} In the case of utilizing global embedding, no additional utterance tokens are added. We use temporal pooling of frame-level output embedding $Y^{\mathcal{S}}$ instead of $U^{\mathcal{S}}$ to compute the loss. 

\subsection{Frame-level Loss} 
Frame-level loss constructs a frame-wise pretext task to learn the context emotion. We only compute the loss on the masked part, which is the common practice for a mask language modeling(MLM) pretext task. The frame-level loss $L_{Frm}$ can be expressed as:

\begin{equation}
L_{Frm} = \frac{1}{M} \sum_{i \in \mathbb{M}} (Y_i^{\mathcal{T}} - Y_i^{\mathcal{S}})^2,
\end{equation}
where $\mathbb{M}$ denotes the index sequence of frame-level output embedding $Y^{\mathcal{S}}$ being masked, and $M$ denotes the total number of tokens being masked. 

\subsection{Online Distillation}
Online distillation is a self-supervised learning strategy for teacher-student learning, where the student network updates parameters by backpropagation and the teacher network updates parameters with an exponentially moving average (EMA)~\citep{grill2020bootstrap}. 
For the student network $\mathcal{S}$, the total loss $L$ for backpropagation is a combination of frame-level loss $L_{Frm}$ and utterance-level loss $L_{Utt}$, donated as:
\begin{equation}
L = L_{Frm} + \alpha L_{Utt},
\end{equation}
with a tunable weight $\alpha$.
For the teacher network $\mathcal{T}$, The parameters $\theta_0^{\mathcal{T}}$ are initialized as the same parameters of the student network $\theta_0^{\mathcal{S}}$, and then are updated with EMA within each mini-batch, donated as:
\begin{equation}
\theta_{t+1}^{\mathcal{T}} = \tau \theta_{t}^{\mathcal{T}} + (1-\tau) \theta_{t+1}^{\mathcal{S}}.
\end{equation}
where $\tau$ is a parameter that increases linearly during pre-training. 
In practice, within each mini-batch the parameters of teacher feature extractor $\mathcal{F}^{\mathcal{T}}$ are copied directly from $\mathcal{F}^{\mathcal{S}}$, while the parameters of teacher backbone network $\mathcal{B}^{\mathcal{T}}$ are updated with EMA from $\mathcal{B}^{\mathcal{T}}$ and $\mathcal{B}^{\mathcal{S}}$.

\section{Experiments Setup}
\subsection{Initial Model}
\label{sec:initial_model}
Different initial models lead to different architectures of feature extractors $\mathcal{F}$, backbone networks $\mathcal{B}$, and initialization parameters $\theta_0$.
Here we adopt two models, data2vec~\footnote{\url{https://dl.fbaipublicfiles.com/fairseq/data2vec/audio_base_ls.pt}} and data2vec 2.0~\footnote{\url{https://dl.fbaipublicfiles.com/fairseq/data2vec2/base_libri.pt}}, both of which have the same feature extractor design but different backbone network designs. 
The feature extractor $\mathcal{F}$ is a 7-layer 1-D convolutional neural network with kernel sizes $(5, 2, 2, 2, 2, 2, 2)$ and strides $(10, 3, 3, 3, 3, 2, 2)$, resulting in 320x downsampling. 
Given the raw audio input $X$ at a 16000 Hz sample rate, the output representations $Z$ are 50 Hz with dimension 512. 
Then a linear projection for dimension transformation from 512 to 768 is applied, followed by the mask operation to construct the input for the backbone network $\mathcal{B}$. 
Here we briefly introduce different backbone networks in data2vec and data2vec 2.0. 
\paragraph{data2vec}
The backbone network $\mathcal{B}$ contains a 5-layer learnable convolutional positional encoding followed by a 12-layer standard Transformer. Each Transformer block is set to 768 model dimension, 3072 bottleneck dimension, and 12 attention heads. 
Finally, a linear projection from 768 to 768 is equipped on the student outputs, the results of which are employed to calculate MLM loss with teacher outputs.
\paragraph{data2vec 2.0}
The data2vec 2.0 model shares the same Transformer architecture with data2vec, except for one more CNN decoder. 
The Transformer encoder only encodes the non-masked parts of downsampled features $Z$, and then the masked parts are complemented with random Gaussian noise before being passed to the CNN decoder, in a MAE-style fashion, to improve efficiency.
The CNN decoder is a 4-layer 1-D convolutional neural network with all kernel sizes set to 7, strides set to 1, and channels set to 384, without downsampling.
A linear projection from 384 to 768 is equipped to compute MLM loss, which works the same way as data2vec.

\subsection{Training Details}
\paragraph{Self-supervised Pre-training}
In the pre-training phase, we train emotion2vec with $262$ hours of unlabeled emotion data shown in Figure~\ref{tab:datasets} with different initial models. 
For the training overhead, The pre-training is conducted on $4$ NVIDIA A10 Tensor Core GPUs, and we simulate $16$ GPUs by setting the update frequency to $4$. We train emotion2vec for $100$ epochs, each of which takes about $37$ minutes. We use a dynamic batchsize, where the maximum number of tokens is $1 \times 10^6$.
For the optimizing strategy, we use Adam with a learning rate of $7.5 \times 10^{-5}$ and a weight decay of $1 \times 10^{-2}$. We train emotion2vec using a cosine learning rate scheduler, with $5\%$ proportion of linear warm-up. 
For the student model, each time step of the input has a probability of $p = 0.5$ to be the start index, and the subsequent $l = 5$ time steps are masked. The hyperparameter  $\alpha$ that controls the loss weight is set to $1$.
For the teacher model, we use the average of the top $k=8$ blocks of the transformer layer outputs for providing the training targets. We apply a linearly increasing strategy for $\tau$ from $\tau_s=0.999$ to $\tau_e=0.99999$ for the teacher parameters exponentially moving average. 

\paragraph{Supervised Fine-tuning}
All model architectures of diverse downstream tasks are designed to be as simple as possible, to demonstrate the representation ability of the pretrained model. 
For the non-sequential task, following the common practice of SUPERB~\cite{yang2021superb}, we use two linear layers with a ReLU activation function sandwiched between them. 
For the sequential task, we use two layers of gated recurrent units (GRU) to make predictions. 

\subsection{Datasets}

\begin{table*}[htbp]
  \centering
  \caption{The datasets at a glance for emotion2vec pre-training and downstream tasks.}
  \label{tab:datasets}
  \resizebox{\linewidth}{!}{
  \begin{tabular}{lllllllll}
    \toprule
    Dataset & Pretrain & Downstream & Source & Emo & Spk & Lang & \#Utts & \#Hours \\
    \midrule
    IEMOCAP~\cite{busso2008iemocap} & \CHECK & \CHECK & Act & 5 & 10 & English & 5531 & 7.0 \\
    MELD~\cite{poria2019meld} & \CHECK & \CHECK & Friends TV & 7 & 407 & English & 13847 & 12.2 \\
    CMU-MOSEI~\cite{zadeh2018multimodal} & \CHECK & \CHECK & YouTube & 7 & 1000 & English & 44977 & 91.9 \\
    MEAD~\cite{wang2020mead} & \CHECK & \CROSS & Act & 8 & 60 & English & 31792 & 37.3 \\
    MSP-Podcast (V1.8)~\cite{martinez2020msp} & \CHECK & \CROSS & Podcast & 8 & 10000+ & English & 72969 & 113.5 \\
    \midrule
    Total  & \CHECK & -- & -- & -- & -- & English & 169053 & 262.0 \\
    \midrule
    CMU-MOSI~\cite{zadeh2016mosi} & \CROSS & \CHECK & YouTube & 7 & 89 & English & 2199 & 2.6 \\
    RAVDESS-Speech~\cite{livingstone2018ryerson} & \CROSS & \CHECK & Act & 8 & 24 & English & 1440 & 1.5 \\
    RAVDESS-Song~\cite{livingstone2018ryerson} & \CROSS & \CHECK & Act & 8 & 23 & English & 1012 & 1.3 \\
    SAVEE~\cite{jackson2014savee} & \CROSS & \CHECK & Act & 7 & 4 & English & 480 & 0.5 \\
    M3ED~\cite{zhao2022m3ed} & \CROSS & \CHECK & TVs & 7 & 626 & Mandarin & 24449 & 9.8 \\
    EmoDB~\cite{burkhardt2005database} & \CROSS & \CHECK & Act & 7 & 10 & German & 535 & 0.4 \\
    EMOVO~\cite{costantini2014emovo} & \CROSS & \CHECK & Act & 7 & 10 & Italian & 588 & 0.5 \\
    CaFE~\cite{gournay2018canadian} & \CROSS & \CHECK & Act & 7 & 12 & French & 936  & 1.2 \\
    SUBESCO~\cite{sultana2021subesco} & \CROSS & \CHECK & Act & 7 & 20 & Bangla & 7000 & 7.8  \\
    ShEMO~\cite{mohamad2019shemo} & \CROSS & \CHECK & Act & 6 & 87 & Persian & 3000 & 3.4 \\
    URDU~\cite{latif2018cross} & \CROSS & \CHECK & Talk shows & 4 & 38 & Urdu & 400 & 0.3 \\
    AESDD~\cite{vryzas2018speech} & \CROSS & \CHECK & Act & 5 & 5 & Greek & 604 & 0.7 \\
    RESD~\cite{aniemore2022resd} & \CROSS & \CHECK & Act & 7 & 200 & Russian & 1396 & 2.3 \\
    \bottomrule
  \end{tabular}
  }
\end{table*}

A summary of the datasets employed in our experiments is presented in Table~\ref{tab:datasets}. There are 18 emotional datasets including 10 different languages: 9 in English, and 1 in Mandarin, Bangla, French, German, Greek, Italian, Persian, Russian, and Urdu. 
For each dataset, it can be categorized in terms of \textit{Pretrain} (\textit{i.e.}, whether used during the pre-training phase), \textit{Downstream} (\textit{i.e.}, whether tested in the downstream task), \textit{Source} (\textit{i.e.}, where samples collected), \textit{Emo} (\textit{i.e.}, number of emotion categories), \textit{Spk} (\textit{i.e.}, number of speakers), \textit{Lang}, (\textit{i.e.}, Language), \textit{\#Utts} (\textit{i.e.}, number of utterances), and \textit{\#Hours}  (\textit{i.e.}, total duration of samples). 
Speech data is extracted from these datasets and uniformly processed into a single channel of 16k Hz. 

In the pretraining phase, we utilize five large-scale English datasets, including IEMOCAP~\cite{busso2008iemocap}, MELD~\cite{poria2019meld}, MEAD~\cite{wang2020mead}, CMU-MOSEI~\cite{zadeh2018multimodal}, and MSP-Podcast~\cite{martinez2020msp}, resulting in a total of 262 hours. The IEMOCAP corpus contains a total of 5 sessions and 10 different speakers, with each session being a conversation of two exclusive speakers. MELD is a multi-party conversational dataset containing about 13,847 utterances from 1,433 dialogues collected from the TV series `\textit{Friends}'. MEAD is a talking-face video corpus featuring 60 actors and actresses talking with 8 different emotions at three different intensity levels. CMU-MOSEI is a multimodal dataset from YouTube for sentiment and emotion analysis in videos. MSP-Podcast is collected from podcast recordings that discuss a variety of topics like politics, sports, and movies. 

Different datasets are used to test different downstream tasks with various languages. 
For main results in Section~\ref{main_results}, we report cross-validation (CV) results on the IEMOCAP dataset. The original labels cover five classes, to be consistent and comparable with previous methods~\cite{ye2023temporal, chen2023dst}, we merge `\textit{excited}' with `\textit{happy}' to better balance the size of each emotion class, resulting in four classes. 
We conduct both leave-one-session-out 5-fold CV and leave-one-speaker-out 10-fold CV. 
Moreover, we report results on MELD under its original split setup, and RAVDESS~\cite{livingstone2018ryerson}, SAVEE~\cite{jackson2014savee} datasets under a random leave-one-out 10-fold CV setup, which implies at each fold, all samples within the dataset are randomly split into 80\%, 10\%, and 10\% samples in training, validation, and testing sets. 
Among them, speech in RAVDESS and SAVEE datasets is not seen in the pre-training stage, which demonstrates the generalization of the proposed model on out-of-domain corpora. 

For language generalization task in Section~\ref{lang_generalization}, we report CV results for 9 out-of-domain datasets, including 1 in Mandarin (M3ED~\cite{zhao2022m3ed}), Bangla (SUBESCO~\cite{sultana2021subesco}), French (CaFE~\cite{gournay2018canadian}), German (EmoDB~\cite{burkhardt2005database}), Greek (AESDD~\cite{vryzas2018speech}), Italian (EMOVO~\cite{costantini2014emovo}), Persian (ShEMO~\cite{mohamad2019shemo}), Russain (RESD~\cite{aniemore2022resd}), and Urdu (URDU~\cite{latif2018cross}). 
If not specified, language generalization results are obtained using the random leave-one-out 10-fold CV as we mentioned above unless the dataset provides a set partition. Such as the RESD dataset, we follow its original split setup with 280 testing samples and 1116 training samples. Additionally, we allocate 10\% from the training samples for validation and others for training. 

For task generalization task in Section~\ref{task_generalization}. We tested other speech emotion tasks, including song emotion recognition, emotion prediction in conversation, and sentiment analysis, on RAVDESS-Song~\cite{livingstone2018ryerson}, IEMOCAP and CMU-MOSI~\cite{zadeh2016mosi} \& CMU-MOSEI~\cite{zadeh2018multimodal}. 
For song emotion recognition and emotion prediction in conversation, we report CV results. 
For sentiment analysis, we report results with its original split setup. 
To be comparable with previous work, the experimental setup varies according to the specific task.

\section{Results}
\subsection{Evaluation Metrics}
We apply commonly used evaluation metrics, weighted accuracy (WA), unweighted accuracy (UA), and weighted average F1 (WF1), to evaluate the performance of speech emotion tasks. 
WA corresponds to the overall accuracy and UA corresponds to the average class-wise accuracy. WF1 is a comprehensive evaluation, especially for the situation of sample imbalance. 

\subsection{Main Results}
\label{main_results}
The results are shown in Table~\ref{tab:SSL Models on SUPERB}, where we compare different SSL pre-trained models on the IEMOCAP dataset, as well as larger-scale pre-trained models, and the latest specialist models designed for SER tasks. 
We follow the evaluation of SUPERB~\cite{yang2021superb}, freezing the pre-trained model and training downstream linear layers with the hidden dimensional set to 256. 
As can be seen from the table, emotion2vec outperforms all existing SSL pre-trained models, across all base models with similar parameters and large models with greater parameters. 
Compared with Versper-12, an SER model obtained by distillation from WavLM-large, emotion2vec works better with fewer parameters. 
TIM-NET~\cite{ye2023temporal}, MSTR~\cite{limulti}, and DST~\cite{chen2023dst} are the latest SER specialist models, respectively, which use different scales of upstream features and downstream networks. 
The proposed emotion2vec model outperforms or performs on par with these models with only linear layers, while their downstream networks have 2x, 135x, and 114x more parameters than emotion2vec, respectively. 
We provide the results of leave-one-session-out five-fold cross-validation and leave-one-speaker-out ten-fold cross-validation for reference. 

We also conduct experiments on other mainstream English datasets to prove the generalization of emotion2vec in Table~\ref{tab:english}.  
MELD is a noisy dataset used to test the SER performance of the model in complex environments. RAVDESS and SAVEE are out-of-domain datasets with respective recording environments. 
Experimental results show that emotion2vec exhibits state-of-the-art performance on different datasets in different environments. 

\begin{table*}[htbp]
\centering
\caption{SER task performance of different SSL pre-trained models on the IEMOCAP dataset. The setting of the downstream models follows SUPERB~\cite{yang2021superb} to use linear layers to test the representation ability of different upstream models. 
``LS-960" means LibriSpeech 960 hours, ``LL-60k" means LibriLight 60k hours, and ``Mix-94k" means 94k hours of data including LibriLight, VoxPopuli, and GigaSpeech. For emotion data, ``LSED-206" means LSED 206 hours, and ``Emo-262" refers to the 262 hours of pre-training data in Table \ref{tab:datasets}. 
Models are tested using leave-one-session-out five-fold cross-validation with 20\% from the training set used as the validation set for each session. Models with \textbf{\underline{underline}} are leave-one-speaker-out ten-fold cross-validation with 8 speakers for training, 1 speaker for validation, and 1 speaker for testing within each fold. Models with \textbf{*} imply the same fold for both validation and testing, for a fair comparison as some work uses this principle. We also compare with \textcolor{gray}{larger-scale pre-trained models} and \textcolor{gray}{the latest specialist models} designed for SER tasks.}
\label{tab:SSL Models on SUPERB}
\resizebox{1\linewidth}{!}{
\begin{tabular}{lcccccc}
\hline
\textbf{Model} & \textbf{Pre-training Corpus} &\textbf{Upstream} &\textbf{\#Upstream Params} &\textbf{Downstream} & \textbf{\#Downstream Params} &\textbf{WA(\%) $\uparrow$} \\
\hline
\multicolumn{7}{l}{\textbf{\textit{Self-supervised Model}}} \\
\hline
\multicolumn{7}{l}{\textit{small size}} \\
wav2vec~\citep{schneider2019wav2vec}  & \multirow{2}{*}{LS-960}   & \multirow{2}{*}{Proposed}  & 32.54M & \multirow{2}{*}{Linear}    &   0.13M      & 59.79        \\
vq-wav2vec~\citep{baevski2019vq}  &    & & 34.15M &   &  0.20M   & 58.24      \\ 
\hdashline
\multicolumn{7}{l}{\textit{base size}} \\
wav2vec 2.0~\citep{baevski2020wav2vec}  & LS-960  & \multirow{12}{*}{Proposed} & 95.04M &  \multirow{12}{*}{Linear}    &   0.20M        & 63.43             \\
HuBERT~\citep{hsu2021hubert}      & LS-960       & & 94.68M &   &   0.20M   & 64.92         \\
WavLM~\citep{chen2022wavlm}        & LS-960    & & 94.70M &   &  0.20M  & 65.94       \\
WavLM+~\citep{chen2022wavlm}       & Mix-94k     & & 94.70M &  &    0.20M    & 67.98          \\
data2vec~\citep{baevski2022data2vec}  & LS-960 &  & 93.75M   &   &    0.20M      &   67.38      \\ 
data2vec 2.0~\citep{baevski2023efficient}   & LS-960      & & 93.78M &  &   0.20M      &  68.58              \\
Vesper-4~\citep{chen2023vesper}    & Mix-94k + LSED-206  & & 63.52 M &  &   0.26M   & 68.40   \\
Vesper-12~\citep{chen2023vesper}   & Mix-94k + LSED-206  &  & 164.29 M   &   &     0.26M    & 70.70 \\
\textbf{emotion2vec}     & LS-960 + Emo-262        & & 93.79M &  &      0.20M      &   \textbf{71.79}          \\
\textbf{emotion2vec*}     & LS-960 + Emo-262                     & & 93.79M &  &      0.20M      &   \textbf{74.48}          \\
\textbf{\underline{emotion2vec}}     & LS-960 + Emo-262        & & 93.79M &  &      0.20M      &   \textbf{72.94}         \\
\textbf{\underline{emotion2vec*}}     & LS-960 + Emo-262                     & & 93.79M &  &      0.20M      &   \textbf{77.64}          \\
\hdashline
\multicolumn{7}{l}{\textcolor{gray}{\textit{large size}}} \\
\textcolor{gray}{wav2vec 2.0~\citep{baevski2020wav2vec}}   &  \textcolor{gray}{LL-60k}    &  \multirow{3}{*}{\textcolor{gray}{Proposed}} & \textcolor{gray}{317.38M} & \multirow{3}{*}{\textcolor{gray}{Linear}}     & \multirow{3}{*}{\textcolor{gray}{0.26M}}    & \textcolor{gray}{65.64}            \\
\textcolor{gray}{HuBERT~\citep{hsu2021hubert}}             &  \textcolor{gray}{LL-60k} & & \textcolor{gray}{316.61M} &  &     & \textcolor{gray}{67.62}         \\
\textcolor{gray}{WavLM~\citep{chen2022wavlm}}            &  \textcolor{gray}{Mix-94k}  & & \textcolor{gray}{316.62M} &    &      & \textcolor{gray}{70.03}     \\
\hline
\multicolumn{7}{l}{\textcolor{gray}{\textbf{\textit{Supervised Model}}}} \\
\hline
\textcolor{gray}{TIM-Net~\citep{ye2023temporal}}     & \multirow{3}{*}{\textcolor{gray}{-}}   & \textcolor{gray}{MFCC} & \textcolor{gray}{-} & \textcolor{gray}{CNN(TIM-Net)} &  \textcolor{gray}{0.40M}    & \textcolor{gray}{68.29}     \\
\textcolor{gray}{MSTR~\citep{limulti}}     &    & \textcolor{gray}{HuBERT-large} &  \textcolor{gray}{316.61M} & \textcolor{gray}{Transformer(MSTR)} &  \textcolor{gray}{27.00M}   & \textcolor{gray}{70.03}     \\
\textcolor{gray}{DST~\citep{chen2023dst}}     &    & \textcolor{gray}{WavLM-large} &  \textcolor{gray}{316.62M} & \textcolor{gray}{Transformer(DST)}   &   \textcolor{gray}{22.78M}    & \textcolor{gray}{71.80}     \\
\hline
\end{tabular}
}
\end{table*}

\begin{table*}[htbp]
  \centering
  \caption{emotion2vec performance on mainstream English datasets. }
  \label{tab:english}
  \resizebox{\linewidth}{!}{
  \begin{tabular}{l|ccc|ccc|ccc}
  \hline
   \multirow{2}{*}{\textbf{Model}} &  \textbf{WA(\%) $\uparrow$}   & \textbf{UA(\%) $\uparrow$}  & \textbf{WF1(\%) $\uparrow$}  & \textbf{WA(\%) $\uparrow$}   & \textbf{UA(\%) $\uparrow$}  & \textbf{WF1(\%) $\uparrow$}  & \textbf{WA(\%) $\uparrow$} & \textbf{UA(\%) $\uparrow$}  & \textbf{WF1(\%) $\uparrow$}  \\
  \cline{2-10}
   & \multicolumn{3}{c|}{\textbf{MELD}}   & \multicolumn{3}{c|}{\textbf{RAVDESS}}  & \multicolumn{3}{c}{\textbf{SAVEE}} \\
  \hline
    WavLM-base	& 46.95	& 16.34	& 35.16	& 37.01	& 37.11	& 36.08  & 42.08 & 38.46 & 38.93  \\
    WavLM-base+ & 43.78	& 16.75	& 34.60	& 38.89	& 38.40 & 37.75 & 43.54 & 39.27 & 42.19 \\
    data2vec	& 45.75	& 24.98	& 43.59	& 69.58	& 69.70	& 69.25  & 82.50	& 82.26	& 82.37 \\
    data2vec 2.0& 48.92	& 26.10	& 45.80	& 81.04	& 80.80 & 80.97 & 83.13	& \textbf{82.94}	& 83.03  \\
    \textbf{emotion2vec}	& \textbf{51.88}	& \textbf{28.03}	& \textbf{48.70}	& \textbf{82.43}	& \textbf{82.86}	& \textbf{82.39} & \textbf{84.38}	& 82.30	& \textbf{84.45} \\
  \hline
    
  \end{tabular}
  }
\end{table*}

\begin{table*}[htbp]

  \centering
  \caption{emotion2vec performance on datasets of other languages. }
  \resizebox{\linewidth}{!}{
  \begin{tabular}{l|ccc|ccc|ccc}
  \hline
   \multirow{2}{*}{\textbf{Model}} &  \textbf{WA(\%) $\uparrow$}   & \textbf{UA(\%) $\uparrow$}  & \textbf{WF1(\%) $\uparrow$}  & \textbf{WA(\%) $\uparrow$}   & \textbf{UA(\%) $\uparrow$}  & \textbf{WF1(\%) $\uparrow$}  & \textbf{WF1(\%) $\uparrow$} & \textbf{UA(\%) $\uparrow$}  & \textbf{WF1(\%) $\uparrow$}\\
  \cline{2-10}
   & \multicolumn{3}{c|}{\textbf{AESD (Gr)}}   & \multicolumn{3}{c|}{\textbf{CAFE (Fr)}}  & \multicolumn{3}{c}{\textbf{RESD (Ru)}} \\
  \hline
    WavLM-base	& 55.33	& 55.50	& 54.86	& 31.61	& 32.02	& 30.88  & 56.17 & 56.17 & 55.69 \\
    WavLM-base+ & 53.83	& 54.41	& 52.48	& 31.40	& 33.39 & 30.40 & 55.00 & 55.19 & 55.08 \\
    data2vec	& 56.67	& 57.26	& 56.57	& 57.10	& 57.68	& 57.36  & 49.42	& 49.77	& 48.97 \\
    data2vec 2.0& 71.33	& 70.20	& 70.93	& 71.51	& 72.98 & 71.50 & 64.08	& 64.33	& 64.17 \\
    \textbf{emotion2vec}	& \textbf{72.33}	& \textbf{72.27}	& \textbf{71.57}	& \textbf{74.52}	& \textbf{75.26}	& \textbf{74.53} & \textbf{64.75}	& \textbf{65.04}	& \textbf{64.53} \\
  \hline
   \textbf{Model} & \multicolumn{3}{c|}{\textbf{EmoDB (De)}}   & \multicolumn{3}{c|}{\textbf{EMOVO (It)}}  & \multicolumn{3}{c}{\textbf{M3ED (Zh)}} \\
  \hline
    WavLM-base	& 59.06	& 55.32	& 58.96	& 40.17	& 40.34	& 37.36  & 44.03 & 18.90 & 34.50 \\
    WavLM-base+ & 65.66	& 64.60	& 64.83	& 40.34	& 41.98 & 40.11 & 45.09 & 20.18 & 36.49 \\
    data2vec	& 67.17	& 64.81	& 66.52	& 51.21	& 51.97	& 49.82  & 44.44	& 21.10	& 37.77\\
    data2vec 2.0& 83.77	& 83.07	& 83.93	& 60.69	& 61.27 & 60.79 & 47.50	& 24.12	& 41.74 \\
    \textbf{emotion2vec}	& \textbf{84.34}	& \textbf{84.85}	& \textbf{84.32}	&	\textbf{61.21}	& \textbf{62.97}	& \textbf{60.89}  & \textbf{49.15}	& \textbf{26.98}	& \textbf{44.38} \\
  \hline
   \textbf{Model} & \multicolumn{3}{c|}{\textbf{SUBESCO (Bn)}}   & \multicolumn{3}{c|}{\textbf{ShEMO (Fa)}}  & \multicolumn{3}{c}{\textbf{URDU (Ur)}} \\
  \hline
    WavLM-base	& 54.50	& 54.77	& 53.96	& 67.27	& 46.60	& 65.63 & 71.00 & 70.25 & 70.82 \\
    WavLM-base+ & 54.73	& 54.69	& 54.59	& 66.73	& 44.29	& 65.12 & 67.25 & 68.68 & 67.47 \\
    data2vec	& 78.29	& 78.25	& 78.21	& 70.80	& 53.96	& 69.84  & 71.75 & 72.67 & 71.83 \\
    data2vec 2.0& 87.91	& 87.95	& 87.90	& 77.90	& 62.03 & 76.96 & 77.50 & 78.42 & 77.12 \\
    \textbf{emotion2vec}	& \textbf{90.91}	& \textbf{90.96}	& \textbf{90.91}	& \textbf{79.97}	& \textbf{66.04}	& \textbf{79.56} & \textbf{81.50} & \textbf{81.87 }& \textbf{81.60} \\
  \hline
    
  \end{tabular}
  }
  \label{tab:language}
\end{table*}

\begin{table*}[htbp]
  \centering
  \caption{emotion2vec performance of the song emotion recognition task on the RAVDESS-Song dataset.}
  \label{tab:song}
  \resizebox{0.8\linewidth}{!}{
  \begin{tabular}{lccccc}
  \hline
  \textbf{Model}  &  \textbf{Upstream}  & \textbf{Downstream} & \textbf{WA(\%) $\uparrow$}   & \textbf{UA(\%) $\uparrow$}  & \textbf{WF1(\%) $\uparrow$} \\
  \hline
  \multicolumn{6}{l}{\textbf{\textit{Self-supervised Model}}} \\
  \hline
  WavLM-base   &  Freeze & \multirow{9}{*}{Linear} &  52.3	& 52.4	   & 52.1 \\
  WavLM-base+  &  Freeze &                 & 54.9	  &  53.9	 & 54.2   \\
  data2vec &  Freeze    &                 & 63.8	  &  64.1	 & 63.4   \\
  data2vec 2.0 &  Freeze&                 & 73.0	  &  74.6	 & 72.7   \\
  L$^3$-Net~\citep{koh2021comparison} &  Freeze&                 & 71.0	  &  -	 & -   \\
  SpecMAE~\citep{sadok2023vector} &  Finetune &                 & 54.5	  &  -	 & 53.9   \\
  VQ-MAE-S (Patch-tf)~\citep{sadok2023vector}  &  Finetune &                 & 84.0	  &  -	 &  84.0   \\
  VQ-MAE-S (Frame)~\citep{sadok2023vector} &  Finetune &                 & 84.2	  &  -	 &  84.3   \\
  \textbf{emotion2vec}   &  Freeze  &           & \textbf{85.0} & \textbf{85.2} &	\textbf{84.8} \\
  \hline
  \multicolumn{6}{l}{\textbf{\textit{\textcolor{gray}{Specialist Model}}}} \\
  \hline
  \textcolor{gray}{VQ-MAE-S (Patch-tf)~\citep{sadok2023vector}} & \multirow{2}{*}{\textcolor{gray}{Finetune}} &  \multirow{2}{*}{\textcolor{gray}{Query2Emo}}           & \textcolor{gray}{83.7}    &  \textcolor{gray}{-}    &  \textcolor{gray}{83.4} \\
  \textcolor{gray}{VQ-MAE-S (Frame)~\citep{sadok2023vector}}  &     &             & \textcolor{gray}{85.8}    &  \textcolor{gray}{-}    & \textcolor{gray}{85.7} \\
  \hline
  \end{tabular}
  }
\end{table*}

\subsection{Language generalization}
\label{lang_generalization}
Given the various languages, the SER datasets exhibit notable domain shifts. The generalization of the model to unseen language is critically important for SER. We validate the generalization of emotion2vec and other baselines on the out-of-domain language SER datasets. 
We follow the evaluation of SUPERB~\cite{yang2021superb}, freezing the pre-trained models and training downstream linear layers with the hidden dimensional set to 256, where the WavLM-base, WavLM-base+, data2vec, dat2vec 2.0, and emotion2vec are our implementations following the practice above. 
As shown in Table~\ref{tab:language}, emotion2vec outperforms all the SSL baseline methods on the 9 different lingual datasets in terms of WA, UA, and WF1. 
These results demonstrate that emotion2vec captures the emotion patterns across languages and shows state-of-the-art performance. 

\subsection{Task generalization}
\label{task_generalization}
In order to verify the generalization of the model, in addition to speech emotion recognition, we tested other speech emotion tasks, including song emotion recognition, emotion prediction in conversation, and sentiment analysis. 

\paragraph{Song Emotion Recognition} 
\label{song_ser}
Song emotion recognition is a sub-task of music emotion recognition (MER), which aims to identify the emotion expressed in a singing voice. 
Following common practice, we perform five-fold cross-validation with randomly shuffled data, and remain one fold unseen during each training, to demonstrate the generalization of the features. 
WavLM-base, WavLM-base+, data2vec, dat2vec 2.0, and emotion2vec are our implementations, following the practice above. The results for L$^3$-NET, SpecMAE, and VQ-MAE-S are taken from their papers. 
As shown in Table~\ref{tab:song}, emotion2vec outperforms all known SSL models even without finetuning the model in the song emotion recognition task. 

\paragraph{Emotion Prediction in Conversation} 
Emotion prediction in conversation (EPC) refers to predicting the future emotional state of a specific speaker based on historical conversation information. 
We reproduce \citeposs{shi2023emotion} method except that the speech features are obtained from our proposed emotion2vec. 
Briefly, the model employs several GRUs with a hierarchical structure for emotion prediction. 
Each prediction takes the previous $6$ turns of the dialogue, in which one speaker can say multiple utterances in each turn. 
The network dimensions, hyperparameters, and training strategies are kept the same as the reference implementation with leave-one-speaker-out 10-fold cross-validation.
For the speech modality, the input is 768-dimensional emotion2vec features.
For the text modality, the input is 378-dimensional BERT~\citep{kenton2019bert} features. 
For the speech + text multimodal, the input is a concatenation of emotion2vec features and BERT features, which also remains the same as the reference implementation. 
As shown in Table~\ref{tab:prediction}, with speech features replaced with emotion2vec in the EPC task, there are performance gains in both speech single modality and speech-text multi-modality. 

\begin{table}[htbp]
  \centering
  \caption{emotion2vec performance of emotion prediction in conversation on the IEMOCAP dataset.}
  \label{tab:prediction}
  \resizebox{\linewidth}{!}{
  \begin{tabular}{llcc}
  \hline
  \textbf{Modality}  & \textbf{Model}   & \textbf{UAR(\%) $\uparrow$}  & \textbf{MacroF1(\%) $\uparrow$} \\
  \multirow{4}{*}{Speech}  &  \citeposs{noroozi2017speech}	& 56.78	   & 55.11 \\
                          &  \citeposs{shi2020dimensional}	  &  61.98	 & 60.21   \\
                           &  \citeposs{shi2023emotion}	  &  65.01	 & 65.91   \\
                          & \textbf{emotion2vec} & 	\textbf{77.19} &	\textbf{76.71} \\
  \hline
  \multirow{3}{*}{Text (for reference)}  &  \citeposs{noroozi2017speech}	&   71.19	   & 70.65 \\
                          &  \citeposs{shi2020dimensional}	  &  74.94	 & 74.54   \\
                       &  \citeposs{shi2023emotion}	  &  77.30	 & 76.67   \\
  \hline
  \multirow{4}{*}{Speech + Text}  &  \citeposs{noroozi2017speech}	&  74.61	   & 73.62 \\
                          &  \citeposs{shi2020dimensional}	  &  76.31	 & 75.50   \\
                       &  \citeposs{shi2023emotion}	  &  80.18	 & 80.01   \\
                         & \textbf{emotion2vec} & 	\textbf{81.68} &	\textbf{80.75} \\
  \hline
  \end{tabular}
  }
\end{table}

\paragraph{Sentiment Analysis} 
Sentiment analysis is the task of analyzing text or speech to determine whether the affective state conveyed is positive, negative, or neutral. 
Following \citeposs{lian2023mer} practice, we eliminate the neutral sentiment and perform the binary classification task on the standard training/validation/test set of CMU-MOSI~\cite{zadeh2016mosi} and CMU-MOSEI~\cite{zadeh2018multimodal}, respectively. 
Also in line with \citeposs{lian2023mer} practice, we utilize the mean of the last four layers' features of the pretrained model, to train the downstream linear layers. 
As shown in Table~\ref{tab:sentiment}, emotion2vec outperforms pretrained data2vec and WavLM with self-supervised learning and pretrained Whisper Encoder~\cite{radford2023robust} with supervised learning utilizing ASR task. 

\begin{table}[htbp]
  \centering
  \caption{emotion2vec performance of sentiment analysis on CMU-MOSI and CMU-MOSEI datasets.}
  \label{tab:sentiment}
  \resizebox{0.8\linewidth}{!}{
  \begin{tabular}{lcc}
  \hline
  \multirow{2}{*}{\textbf{Model}}  & \multicolumn{2}{c}{\textbf{WF1(\%) $\uparrow$}} \\
  \cline{2-3}
    & \textbf{CMU-MOSI}   & \textbf{CMU-MOSEI} \\
  \hline
  \multicolumn{3}{l}{\textbf{\textit{Base Size}}} \\
  \hline
  data2vec & 65.06 & 72.79 \\
  WavLM & 62.36 & 73.47 \\
  Whisper Encoder & 65.41 & 74.75 \\
  \textbf{emotion2vec} &  \textbf{69.16} & \textbf{76.56} \\
  \hline
  \multicolumn{3}{l}{\textcolor{gray}{\textbf{\textit{Large Size}}}} \\
  \hline
  \textcolor{gray}{WavLM} & \textcolor{gray}{68.27} &  \textcolor{gray}{77.66} \\
  \textcolor{gray}{Whisper Encoder} & \textcolor{gray}{64.93} & \textcolor{gray}{76.46} \\
  \hline
  \end{tabular}
  }
\end{table}

\subsection{Ablation Study}
If not specified, results are obtained using the standard leave-one-session-out 5-fold cross-validation on the IEMOCAP dataset. 

\paragraph{Initialization Method}
In this experiment, we explore the impact of initialization methods on performance. 
data2vec and data2vec 2.0 are two representative models trained with online distillation, both of which are pre-trained on Librispeech 960 hours. 
As shown in Table~\ref{tab:init}, initializing with a pre-trained model would be better than the cold start method. Model initializing with data2vec 2.0 performs better than the one initializing with data2vec. 

\begin{table}[htbp]
  \centering
  \caption{Ablation study with initialization methods.}
  \label{tab:init}
  \resizebox{0.9\linewidth}{!}{
  \begin{tabular}{lccc}
  \hline
  \textbf{Initialization}  & \textbf{WA(\%) $\uparrow$}   & \textbf{UA(\%) $\uparrow$}  & \textbf{WF1(\%) $\uparrow$} \\
  Cold Start  &  61.34	& 62.71	   & 61.19 \\
  data2vec   &  70.2	  &  70.93	 & 70.11   \\
  data2vec 2.0  & \textbf{71.79} & 	\textbf{72.69} &	\textbf{71.80} \\
  \hline
  \end{tabular}
  }
\end{table}

\paragraph{Training Loss}
In this experiment, we explore the impact on performance of different combinations of loss when pre-training, and different features when training downstream models. 
As shown in Table~\ref{tab:loss_type}, if only utterance-level loss is adopted during pre-training, the model almost does not work. 
If pre-trained with frame-level loss, the model obtains reasonable results whether utterance-level loss exists or not.
When pre-trained with both utterance-level loss and frame-level loss, the model achieves good results. 
We also try to concatenate utterance embeddings and frame embeddings when training downstream models, where we obtain similar results as frame embeddings only. 

\begin{table}[htbp]
  \centering
  \caption{Ablation study with different loss.}
  \label{tab:loss_type}
  \resizebox{\linewidth}{!}{
  \begin{tabular}{cccccc}
  \hline
  \textbf{Frm Loss}  &  \textbf{Utt Loss}  &  \textbf{Downstream}  & \textbf{WA(\%) $\uparrow$}   & \textbf{UA(\%) $\uparrow$}  & \textbf{WF1(\%) $\uparrow$} \\
  \CROSS  &  \CHECK   & Utt & 28.96 & 25.0   & 13.13   \\
  \CHECK  &  \CROSS	& Frm	&  70.85	&  71.61 &	70.71 \\
  \CHECK  &  \CHECK   & Utt &  62.77 & 63.53   & 62.43   \\
  \textbf{\CHECK}  &  \textbf{\CHECK}   & Frm  &  \textbf{71.79} & 	\textbf{72.69} &	\textbf{71.80}   \\
  \CHECK  &  \CHECK   & Utt $\oplus$ Frm   & 71.37  &  \textbf{72.69}  &	71.29   \\
  \hline
  \end{tabular}
  }
\end{table}

\paragraph{Utterance-level Loss Method}
In this experiment, we compare the impact of different types of utterance-level loss proposed in Section~\ref{sec:utterance-level loss} on performance. 
As shown in Table~\ref{tab:utterance-level loss}, chunk embedding chosen to compute utterance-level loss performs the best.

\begin{table}[htbp]
  \centering
  \caption{Ablation study with the methods for utterance-level loss.}
  \label{tab:utterance-level loss}
  \resizebox{0.9\linewidth}{!}{
  \begin{tabular}{lccc}
  \hline
  \textbf{Utt Loss Method}  & \textbf{WA(\%) $\uparrow$}   & \textbf{UA(\%) $\uparrow$}  & \textbf{WF1(\%) $\uparrow$} \\
  Token  &  70.46	& 71.07	   & 70.33 \\
  Chunk  &  \textbf{71.79} & 	\textbf{72.69} & \textbf{71.80}   \\
  Global  & 70.30 & 71.52 &	70.18 \\
  \hline
  \end{tabular}
  }
\end{table}

\paragraph{Utterance-level Loss Weight}
In this experiment, we compare the impact of utterance-level loss weight on performance. 
As shown in Table~\ref{tab:weight}, weighting utterance-level loss and frame-level loss with a ratio of 1:1 works best. 

\begin{table}[htbp]
  \centering
  \caption{Ablation study with the weight for utterance-level loss.}
  \label{tab:weight}
  \resizebox{0.9\linewidth}{!}{
  \begin{tabular}{lccc}
  \hline
  \textbf{Utt Loss $\alpha$}  & \textbf{WA(\%) $\uparrow$}   & \textbf{UA(\%) $\uparrow$}  & \textbf{WF1(\%) $\uparrow$} \\
  0 & 70.85	&  71.61 &	70.71 \\
  0.1   &  71.06	  &  72.16	 & 71.08   \\
  1  & \textbf{72.14} & 	\textbf{72.84} &	\textbf{72.13} \\
  10 & 70.58 & 71.37 & 70.61 \\
  \hline
  \end{tabular}
  }
\end{table}

\begin{figure*}[htbp]
  \centering
  \includegraphics[width=1\textwidth]{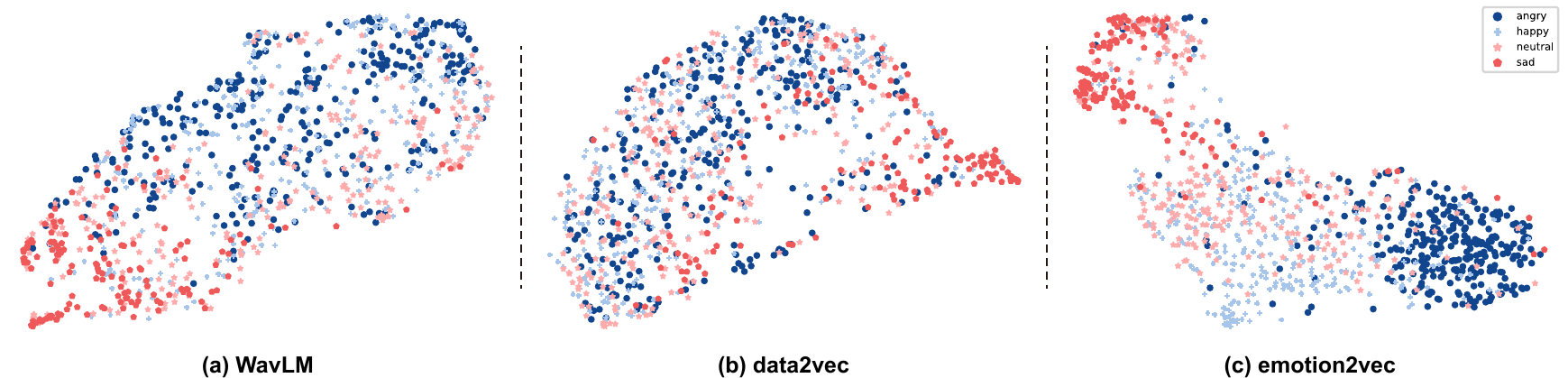}
  \caption{UMAP visualizations of learned features on downstream SER task from WavLM, data2vec, and emotion2vec on the IEMOCAP dataset. \textcolor{red}{Red} and \textcolor{blue}{Blue} tones mean \textcolor{red}{low} and \textcolor{blue}{high} arousal emotional classes, respectively.}
  \label{fig:IEMOCAP}
\end{figure*}

\begin{figure*}[htbp]
  \centering
  \includegraphics[width=1\textwidth]{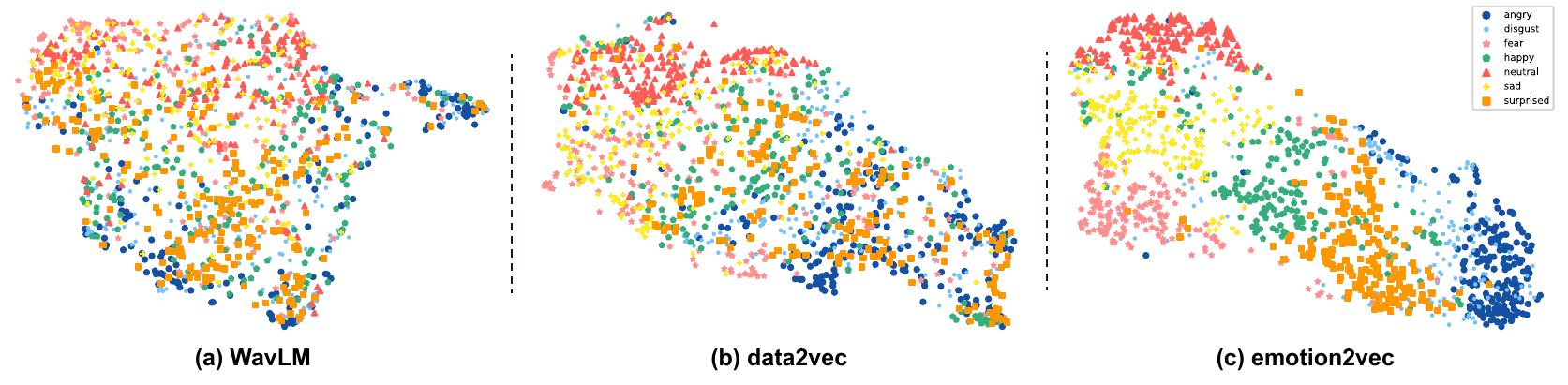}
  \caption{UMAP visualizations of learned features on downstream SER task from WavLM, data2vec, and emotion2vec on the SUBESCO dataset.}
  \label{fig:SUBESCO}
\end{figure*}

\subsection{Visualization}
To investigate the intuitive effect of emotion2vec and other SSL baselines on emotion representation learning, we visualize the representations learned by WavLM, data2vec, and emotion2vec through the UMAP technique~\cite{mcinnes2018umap} in Figure~\ref{fig:IEMOCAP} and Figure~\ref{fig:SUBESCO}. 
We conduct the leave-one-session-out evaluation strategy on IEMOCAP, and the 8:2 hold-out evaluation on SUBESCO, both of which we randomly select 10\% samples from the training set as the validation set. 
Specifically, for a fair comparison, the representations from the first linear layer are visualized after an identical training phase for different SSL models. 

Figure~\ref{fig:IEMOCAP} visualizes different SSL models to represent arousal. In a sense, arousal refers to emotional intensity. 
Figure~\ref{fig:IEMOCAP} (a) and Figure~\ref{fig:IEMOCAP} (b) show heavy overlapping between high and low arousal emotion classes. In contrast, Figure~\ref{fig:IEMOCAP} (c) shows that the high arousal and low arousal representations are clustered receptively, and the feature distribution exhibits a trend transitioning from high arousal to low arousal, which is more reasonable compared to other methods. 
Figure~\ref{fig:SUBESCO} shows the ability of different SSL models to represent discrete emotion classes. 
As Figure~\ref{fig:SUBESCO} (a) and Figure~\ref{fig:SUBESCO} (b) show, WavLM and data2vec suffer from class confusion problems. On the contrary, the features learned by emotion2vec demonstrate a higher intra-class compactness and a much larger inter-class margin. 
The results indicate that emotion2vec provides more class-discriminative and emotion-aware representations to support its superior performance.

\section{Conclusion}
In this paper, we propose emotion2vec, a universal emotion representation model. 
emotion2vec is pre-trained on 262 hours of unlabeled emotion data through self-supervised online distillation, leading to universal emotion representation ability. 
We prove that our strategy of combining utterance-level loss and frame-level loss during emotion pre-training is effective. 
Extensive experiments demonstrate that the proposed emotion2vec has the ability to extract emotion representation across different tasks, languages, and scenarios. 
In the future, we will explore the scaling law of emotion representation models, namely how to provide a better representation with more data and larger parameters. 

\bibliography{custom}

\end{document}